# Evaluating the Effect of Longitudinal Dose and INR Data on Maintenance Warfarin Dose Predictions

Anish Karpurapu[1*], Adam Krekorian[2*], Ye Tian [1*], Leslie M. Collins[2],
Ravi Karra[3], Aaron Franklin[2], Boyla O. Mainsah[2]

[1]Deparment of Biomedical Engineering, Duke University, Durham, NC, USA
[2]Department of Electrical and Computer Engineering, Duke University, Durham, NC, USA
[3]Department of Cardiology, Duke University Medical Center, Durham, NC, USA

**ABSTRACT**

Warfarin, a commonly prescribed drug to prevent blood clots, has a highly variable individual response. Determining a maintenance warfarin dose that achieves a therapeutic blood clotting time, as measured by the international normalized ratio (INR), is crucial in preventing complications, such as excessive bleeding with an overdose or blood clotting with an underdose. Machine learning models are increasingly being used to automate warfarin dosing; usually, an initial dose is predicted with clinical and genotype factors, and this dose is revised after a few days based on previous doses and the current INR. Since the history of prior doses and INR better capture the variability in individual warfarin response, we hypothesized that longitudinal dose response data will improve maintenance dose predictions with dose revision algorithms. To test this hypothesis, we analyzed a dataset from a prospective warfarin dosing clinical trial (the COAG study), which includes demographics, clinical data, warfarin doses and INR measurements over the study period, and maintenance dose when therapeutic INR was achieved. Dose revision algorithms to predict maintenance dose were developed by training various machine learning regression models with clinical factors and previous warfarin doses and INR data as features. Overall, dose revision algorithms with a single dose and INR measurement achieved comparable performance as the baseline dose revision algorithm. In contrast, dose revision algorithms with longitudinal dose and INR data provided statistically significantly better maintenance dose predictions that were much closer to the true maintenance dose and better explained the variability of the maintenance dose. Focusing on the best performing model, gradient boosting, the proportion of ideal estimated dose, i.e., defined as within ±20% of the true dose, increased from the baseline (54.92%) to the gradient boosting model with the single (63.11%) and longitudinal (75.41%) INR feature. More accurate maintenance dose predictions with longitudinal dose response data can potentially achieve therapeutic INR faster, reduce drug-related complications and improve patient outcomes with warfarin therapy.

*Authors contributed equally to this work. This work was funded by the Duke University Phoenix Project.

# 1 INTRODUCTION

Warfarin is a commonly prescribed anticoagulant for blood clotting disorders [1]. Warfarin is taken daily to maintain an individual's blood clotting ability within a therapeutic range, as measured by the *international normalized ratio* (INR), a standard measure of blood thickness [1]. An INR of 1.0 is typical of a healthy individual [2]. For individuals with blood clotting disorders, a therapeutic INR range between 2-3 is desired [2]. The appropriate warfarin dose is difficult to establish because of the narrow therapeutic INR range and the high variability in individual response to warfarin due to differences in warfarin sensitivities across individuals [3]. Too much warfarin can predispose to bleeding events, often necessitating warfarin cessation. Too little warfarin can lead to increased risk for blood clotting, or thrombosis. Both scenarios can lead to prolonged hospitalizations and frequent clinic visits with inadequate dose selection; warfarin is a major cause of drug-related hospitalizations and deaths [4, 5]. Thus, strategies to improve warfarin dosing are critically needed.

A precise warfarin dose is crucial but also challenging to determine clinically. The dose to achieve a stable therapeutic INR, is termed the *maintenance dose*[1, 2]. Calculating the proper maintenance dose involves frequent blood tests to monitor INR for dosage adjustments. The most common warfarin dosing strategy is to initiate with a fixed dose (usually 5 mg/day) and update the dose heuristically to achieve therapeutic INR using physician-guided dose adjustments based on observed INR [1, 2]. A fixed initial dose and heuristic approach to update warfarin dosing often results in fluctuations in INR, which can increase the risk of thrombosis or bleeding. Thus, automated models that accurately predict the maintenance dose of warfarin are needed.

Several research groups have developed multivariate linear and nonlinear machine learning models to predict the maintenance warfarin dose [6] Approaches range from dosing algorithms that incorporate clinical factors, such as age, size, race, smoking status, concurrent medications, and disease condition, to algorithms that consider both clinical and genetic factors, such as CYP2C9 and VKORC1 alleles, which are genetic variants of enzymes that metabolize for warfarin. Usually, an initial dose is predicted with a *dose initiation* algorithm that uses only clinical or clinical and genotype characteristics, e.g.,[7], and the dose is refined after a few days with a *dose revision* algorithm that includes the previous warfarin doses and the current INR as additional variables, e.g., [8]. Dose revision algorithms generally perform better than dose initiation algorithms [6], likely because the former incorporates individual dose response data that are more predictive of the maintenance warfarin dose.

A potential limitation with using only the current INR in a dose revision algorithm is that it may not fully capture differences in warfarin sensitivities across individuals that are better reflected in the dosing history and INR changes over time. Thus, we hypothesized that incorporating additional INR measurements into a dose revision algorithm can more accurately predict the maintenance dose. Here, we investigate whether longitudinal warfarin dose and INR data improves warfarin dose predictions by analyzing a dataset from a prospective warfarin dosing clinical trial.

## 2 METHODS

### 2.1 Dataset

We obtained the Clarification of Optimal Anticoagulation through Genetics (COAG) study dataset through the Biologic Specimen and Data Repository Information Coordinating Center (BioLINCC) repository developed by the National Heart, Lung and Blood Institute (NHLBI) [9]. Briefly, the COAG trial was an NHLBI-funded prospective clinical trial to determine whether a warfarin dosing algorithm based on clinical factors or clinical and genetic factors could better predict the warfarin maintenance dose [10]. In the trial, the initial warfarin dose (mg/day) was calculated using either a clinical or genotype dose initiation algorithm. The dose was then updated after 4-5 days using the relevant revision dose algorithm that also incorporated previous warfarin doses and the current INR.

The COAG BioLINCC dataset had a total of 882 participants. The dataset included demographic characteristics, clinical data, such as indication of warfarin use, concomitant medications, etc., and longitudinal warfarin dose and measured INR data over the study period of 6 months. Demographics and clinical characteristics of the COAG BioLINCC participants are shown in Table 1. The maintenance dose (mg/day) that achieved therapeutic INR was included. Fig. 1 shows the distribution of the maintenance dose.

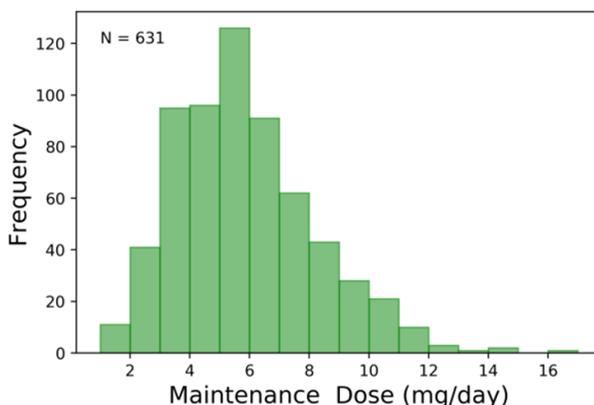

**Figure 1: Distribution of maintenance warfarin doses in the COAG BioLINCC dataset.**

The number of INR measurements and the time interval between measurements were inconsistent across patients, particularly for data collected in the later days of the trial. Thus, we restricted the data for consideration between 7-10 days to minimize differences in sampling times across participants and to be within a clinically relevant time frame for another dose revision [11]. The COAG BioLINCC dataset was pruned to exclude participants that did not achieve a maintenance dose, had no clinical revision dose and no INR data between 7-10 days. After pruning the dataset for completeness, data from 609 participants were available for model development and analysis, Fig. 2.

**Table 1: Characteristics of participants in the COAG BioLINCC dataset.**

| CHARACTERISTIC | TOTAL (N = 882) |
|---|---|
| Age - years, median (IQR) | 58 (46, 69) |
| Male Sex - count (%) | 451 (51) |
| Race - count (%) | |
|     White | 581 (67) |
|     African American | 232 (26) |
|     Hispanic | 53 (6) |
|     Asian | 16 (2) |
| Weight - kg, median (IQR) | 87.0 (74.8, 105.7) |
| Indication for Anticoagulation - number (%) | |
|     Atrial fibrillation or flutter | 187 (21) |
|     PE/DVT | 508 (58) |
|     Other indication | 100 (11) |
|     Multiple indications | 79 (9) |
|     No indication given | 8 (1) |

Data reported as median (25th, 75th percentiles) or count (percentage).
IQR, Interquartile range; PE, Pulmonary Embolism; DVT, Deep vein thrombosis.

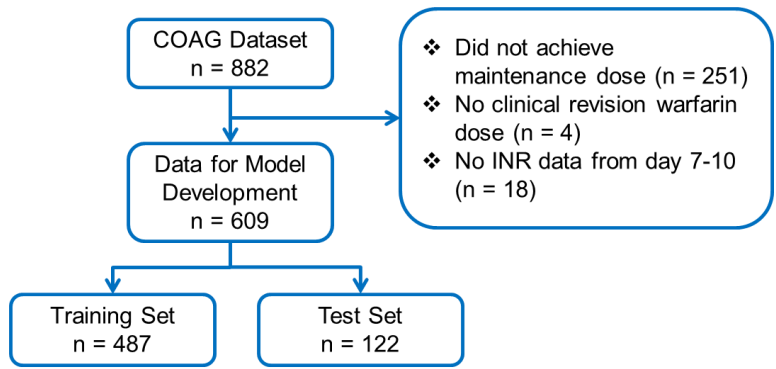

**Figure 2: Flowchart showing how the COAG BioLINCC dataset was pruned for warfarin dose prediction algorithm development.**

### 2.2 Features

We initially focused on clinical dose revision algorithms since genetic testing may not be readily available in every clinical setting; future work will explore genotype dose revision algorithms. Base clinical variables included age, gender, race, and weight. Age was binned by decade. Gender was represented as a binary variable for male and female. Race was represented as a binary variable for each racial category, which included African American, White, Asian, and other. In the clinical dose revision algorithm used in the COAG trial [8, 10], the first INR within 4-5 days from dose initiation and the three previous warfarin doses. To model a similar feature, we added the previous dose and INR measurement within the 7 to 10-day time frame to the base clinical variables, which we will refer to as the *single INR* feature. We then incorporated additional dose response data by using the next INR and the corresponding previous dose, which we will refer to as a *longitudinal INR* feature. The various feature sets are detailed in Table 2.

**Table 2: Feature sets for maintenance warfarin dose predictions.**

| FEATURE SET | VARIABLES |
|---|---|
| **Clinical** | Sex, Race, Age, Weight |
| **Single INR** | Clinical + Dose($t_1 - 1$), INR($t_1$) |
| **Longitudinal INR** | Clinical + Dose($t_1 - 1$), INR($t_1$), Dose($t_1$), INR($t_2$) |

$t_1$ = day of the first INR measurement within 7-10 days from dose initiation.
$t_2$ = day of the next INR measurement after $t_1$.

### 2.3 Model Development

Various machine learning regression models were trained to predict the maintenance warfarin dose using the various feature sets. The regression models included: linear, ridge, Bayesian ridge, decision tree, gradient boosting, and multilayer perceptron. For each model, the dataset was randomly split into 80% training set and 20% test set with equal racial proportions in each set. Ten-fold cross validation was performed during training to optimize model parameters. The trained models were evaluated on the test set. All models were trained and tested in Python with the scikit-learn library.

### 2.4 Performance Measures

Prediction performance was evaluated using mean squared error (MSE), mean absolute error (MAE) and the coefficient of determination ($R^2$) between the actual and predicted maintenance doses. The 95% confidence intervals (CIs) of the MSE and MAE were used to evaluate the degree of closeness between the predicted and true maintenance doses [12]. Statistical significance was evaluated at $p$ <0.05. The percentage of *ideal estimated dose*, defined as a predicted dose within ±20% of the true maintenance dose, was also calculated [13]. Statistical analysis was performed in Python.

### 3 RESULTS

The dose revision algorithms were compared against a baseline algorithm, the clinical dose revision algorithm used in the COAG trial [8, 10]. Fig. 3 shows the root MSE and MAE of the various regression models with the different feature sets. The addition of dose and INR data to the clinical variables substantially improved dose prediction accuracy, confirming the utility of warfarin dose revision with dose response data. Dose revision algorithms with the single INR feature had similar performance as the baseline revision algorithm, except for the gradient boosting and multi-layer perceptron models that performed better. Overall, dose revision algorithms with the longitudinal INR feature predicted maintenance warfarin doses that were the closest to the true maintenance dose.

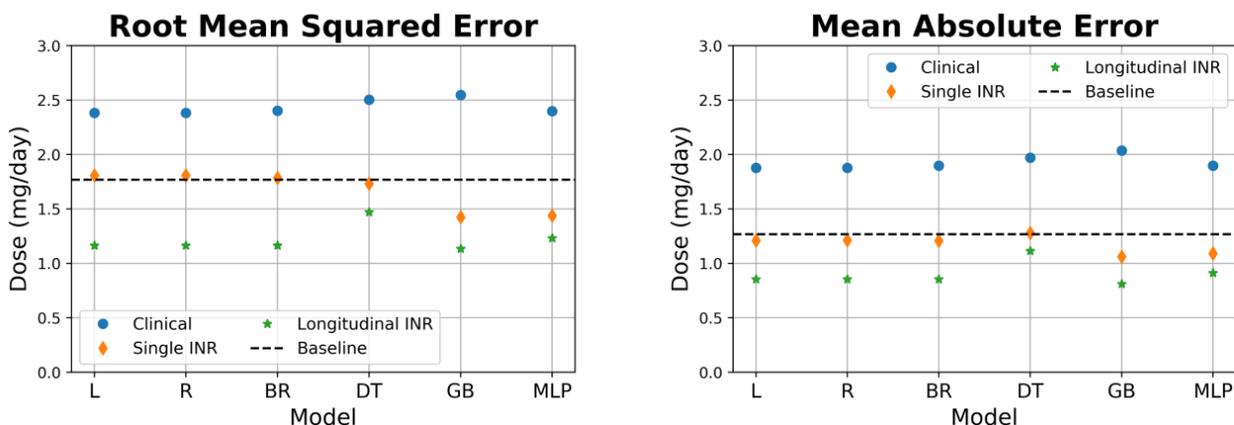

**Figure 3: Root mean squared error and mean absolute error of the maintenance warfarin dose prediction algorithms.** The machine learning models include linear (L), ridge (R), Bayesian ridge (BR), decision tree (DT), gradient boosting (GB) and multilayer perceptron (MLP) regression. The feature sets include clinical, clinical and single INR feature, and clinical and longitudinal INR feature. The baseline model is the clinical dose revision algorithm used in the COAG trial.

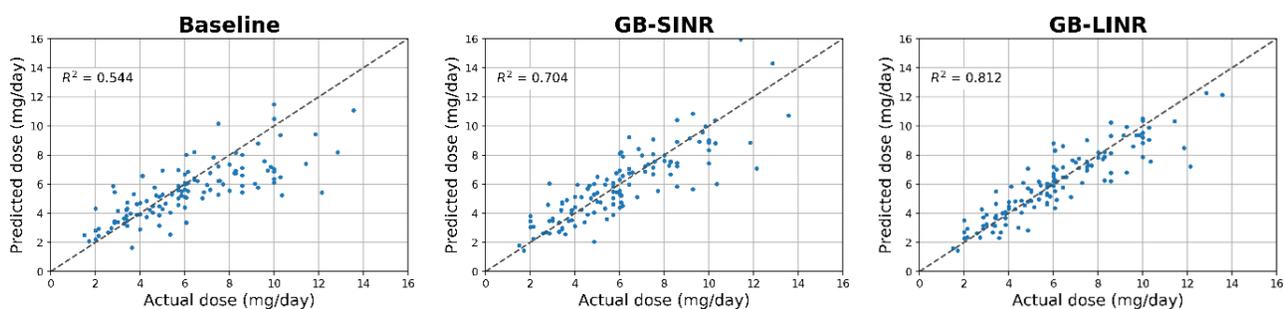

**Table 3: Actual and predicted maintenance warfarin doses for the baseline dose revision algorithms and the gradient boosting (GB) dose revision algorithm with single INR (SINR) and longitudinal INR (LINR) features.** The coefficient of determination ($R^2$) is indicated.

**Table 4: Prediction performance of the baseline dose revision algorithm and gradient boosting (GB) algorithms with single INR (SINR) and longitudinal INR (LINR) features.**

| PERFORMANCE MEASURE | BASELINE | GB-SINR | GB-LINR |
|---|---|---|---|
| MSE (mg$^2$/day$^2$) | 3.13 (2.07, 4.19) | 2.03 (1.33, 2.73) | 1.29 (0.78, 1.79) |
| $\Delta$MSE $^{base}$(mg$^2$/day$^2$) | | 1.10 (0.36, 1.83) | 1.84 (1.02, 2.66) |
| MAE (mg/day) | 1.27 (1.05, 1.49) | 1.06 (0.89, 1.23) | 0.81 (0.67, 0.95) |
| $\Delta$MAE $^{base}$ (mg/day) | | 0.21 (0.02, 0.40) | 0.46 (0.26, 0.65) |
| Coefficient of Determination ($R^2$) | 0.54 | 0.70 | 0.81 |
| Ideal Estimated Dose (%) | 54.9% | 63.11% | 75.41% |

Mean squared error (MSE), mean absolute error (MAE) and 95% confidence intervals (CIs).
$\Delta$[MSE, MAE] $^{base}$ is the difference between baseline and the relevant model.
$\Delta^{base}$ 95% CI without zero denotes a statistically significant improvement from the baseline.

For subsequent analysis, we focus on the overall best performing model, gradient boosting (GB) regression. The prediction performance of the baseline algorithm and the GB algorithms are summarized in Table 4. The predicted and actual doses for the baseline and GB algorithms are shown in Fig. 3. The $R^2$ increased from the baseline algorithm ($R^2$ = 0.54), to GB with single INR (GB-SINR, $R^2$ = 0.70), to GB with longitudinal INR (GB-LINR, $R^2$ = 0.81). The baseline revision algorithm had a MSE of 3.13 [95% CI 2.07, 4.19] mg $^2$/day $^2$ and MAE 1.27 [1.05, 1.49] mg/day. The GB-SINR model had a MSE of 2.03 [1.33, 2.73] mg $^2$/day $^2$ and MAE of 1.06 [0.89, 1.23] mg/day. The GB-LINR had a MSE of 1.29 [0.78, 1.791] mg $^2$/day $^2$ and MAE of 0.81 [0.67, 0.95 mg/day. The 95% confidence interval differences of the MSE and MAE between the baseline algorithm and each GB algorithm were also evaluated; a 95% CI of the difference with no zero value indicates a statistically significant improvement in performance from the baseline [12]. Both GB dose revision algorithms provided dose estimates that were statistically significantly closer to the true maintenance doses when compared to the baseline algorithm; however, the GB-LINR algorithm had statistically significantly better prediction accuracy than the GB-SINR algorithm (Table 4). The proportion of participants with the ideal estimated dose increased from baseline (54.92%) to GB-SINR (63.11%) to GB-LINR (75.41%).

## 4 DISCUSSION

In this work, we investigated the utility of additional longitudinal dose response data to warfarin dose revision. Consistent with our hypothesis, maintenance dose predictions progressively improved with the addition of more warfarin and INR measurements. Models that included longitudinal dose and INR data could explain more of the variability of the maintenance dose and predicted doses that were much closer to the true maintenance dose, with a higher proportion of predicted doses within a clinically acceptable error range. Limitations of this work are the small dataset size and the lack of clinical validation. Future work includes model development using larger and more diverse datasets and clinical validation in a prospective warfarin dosing trial.

In summary, we have shown that incorporating longitudinal dosing history and INR measurments improves maintenance warfarin dose predictions. This work is highly relevant as more accurate maintenance dose prediction algorithms that exploit longitudinal dose response data have the potential to achieve therapeutic INR faster, reduce drug-related complications and improve patients outcomes with warfarin therapy.